\documentclass[conference]{IEEEtran}
\IEEEoverridecommandlockouts
\usepackage{url}
\usepackage{booktabs}       % professional-quality tables
\usepackage{amsfonts}       % blackboard math symbols
\usepackage{nicefrac}       % compact symbols for 1/2, etc.
\usepackage{microtype}      % microtypography

\usepackage{multirow}
\usepackage{array}
\usepackage{verbatim}
\usepackage{caption}
\usepackage{indentfirst}
\usepackage{setspace}
\usepackage{algorithm, listings}
\usepackage{mathtools}
\usepackage{commath}
\usepackage{hhline}
\usepackage{subfigure}

\usepackage{subcaption}
\usepackage{wrapfig}
\usepackage{float}
\usepackage{enumitem}
\usepackage{bm}

\def\BibTeX{{\rm B\kern-.05em{\sc i\kern-.025em b}\kern-.08em
    T\kern-.1667em\lower.7ex\hbox{E}\kern-.125emX}}

\newcommand{\vs}{ \vspace{-5pt} }

\begin{document}

\title{Robust Low-Light Human Pose Estimation through Illumination-Texture Modulation \\

\thanks{Lei Chen is the corresponding author.}

\author{
\IEEEauthorblockN{
 Feng Zhang\IEEEauthorrefmark{2},
 Ze Li\IEEEauthorrefmark{2},
Xiatian Zhu\IEEEauthorrefmark{3},
 Lei Chen\IEEEauthorrefmark{2}}
\IEEEauthorblockA{\IEEEauthorrefmark{2}
\textit{Nanjing University of Posts and Telecommunications, Nanjing, China} \\
\{zhangfengwcy,~lize3329\}@gmail.com, chenlei@njupt.edu.cn}
\IEEEauthorblockA{\IEEEauthorrefmark{3}
\textit{Surrey Institute for People-Centred 
Artificial Intelligence, University of Surrey Guildford, United Kingdom} \\
eddy.zhuxt@gmail.com}
}
}

\maketitle

\begin{abstract}
As critical visual details become obscured, the low visibility and high ISO noise in extremely low-light images pose a significant challenge to human pose estimation.
Current methods fail to provide high-quality representations due to reliance on pixel-level enhancements that compromise semantics and the inability to effectively handle extreme low-light conditions for robust feature learning. 
In this work, we propose a frequency-based framework for low-light human pose estimation, rooted in the "divide-and-conquer" principle. 
Instead of uniformly enhancing the entire image, our method focuses on task-relevant information. 
By applying dynamic illumination correction to the low-frequency components and low-rank denoising to the high-frequency components, we effectively enhance both the semantic and texture information essential for accurate pose estimation. 
As a result, this targeted enhancement method results in robust, high-quality representations, significantly improving pose estimation performance.
Extensive experiments demonstrating its superiority over state-of-the-art methods in various challenging low-light scenarios.

\end{abstract}

\begin{IEEEkeywords}
human pose estimation, low-light image, illuminance correction, low-rank denoising
\end{IEEEkeywords}

\section{Introduction}

Human pose estimation is essential for various applications, such as autonomous driving \cite{zanfir2023hum3dil,sengupta2020mm} and intelligent surveillance \cite{cucchiara2022fine,cormier2022we}.
Yet, the majority of existing research predominantly targets well-lit scenarios \cite{toshev2014deeppose,li2021human,sun2019deep}, neglecting the significant challenges posed by low-light conditions. 
The characteristics of low-light images, such as low contrast, poor visibility, and high ISO noise, significantly impair the performance of existing models, thereby limiting the practical application of human pose estimation in real-world scenarios.

Current research usually tackles low-light challenges by %integrating 
resorting to existing enhancement techniques \cite{guo2016lime, jiang2021enlightengan,guo2020zero, liu2021retinex}, but such workarounds often lead to suboptimal results. They primarily adjust brightness and contrast at the pixel level to meet human visual perception, often compromising semantic information. In addition, some approaches require strictly paired low-light and well-lit images, which are difficult to obtain in practice, limiting their real-world applicability.
While attempts have been made to learn illumination-invariant features \cite{lee2023human,cui2021multitask}, they still depend on paired data and struggle with large illumination differences.
Hashmi et al. \cite{hashmi2023featenhancer} proposed a machine perception-centric approach, using end-to-end training to learn the enhancement of low-light image semantics in a task-driven manner. However, in extreme low-light scenarios, the lack of sufficient visual cues makes it difficult for such methods to extract useful semantic information from images, leading to a significant decline in model performance.

To better enhance the information required for human pose estimation tasks, we advocate a "divide-and-conquer" strategy that decouples and processes task-relevant information instead of uniformly enhancing the entire image. This strategy allows for more effective extraction and enhancement of task-specific information. 
Thus, we introduce a representation learning framework based on frequency decoupling, which applying dynamic illumination correction to low-frequency components and low-rank denoising to high-frequency components, improving semantic and texture information crucial for pose estimation under low-light conditions. 
Specifically, we learn a set of global-local illumination correction curve coefficients to ensure effective and efficient adjustments. The illumination correction is further accelerated by utilizing a Taylor series expansion, which speeds up the computation. Exploiting the property that high-frequency components have inherently low-dimensional structure \cite{schiller2010parallel}, we extend the low-rank approximation theroy \cite{markovsky2012low} to design a multi-scale low-rank denoising module with minimal parameters. 
Furthermore, our method, designed as an efficient module aligned with human pose estimation task, is trained end-to-end with only the task loss, providing robust and high-quality representations that significantly improve the performance of low-light human pose estimation model.

We summarize the contributions as follows:
\begin{enumerate}
\item A efficient enhancement method consists of dynamic illumination correction and low-rank denoising between the frequency decomposition and reconstruction for effective low-light human pose estimation.
\item An effective and robust end-to-end representation learning framework that jointly optimizes low-light enhancement and human pose estimation.
\item Experiments demonstrate the significant performance improvements for low-light human pose estimation.
\end{enumerate}

\section{Methods}

\subsection{Overview}

\begin{figure*}[htbp]
\centering {\includegraphics[width=0.8\textwidth]{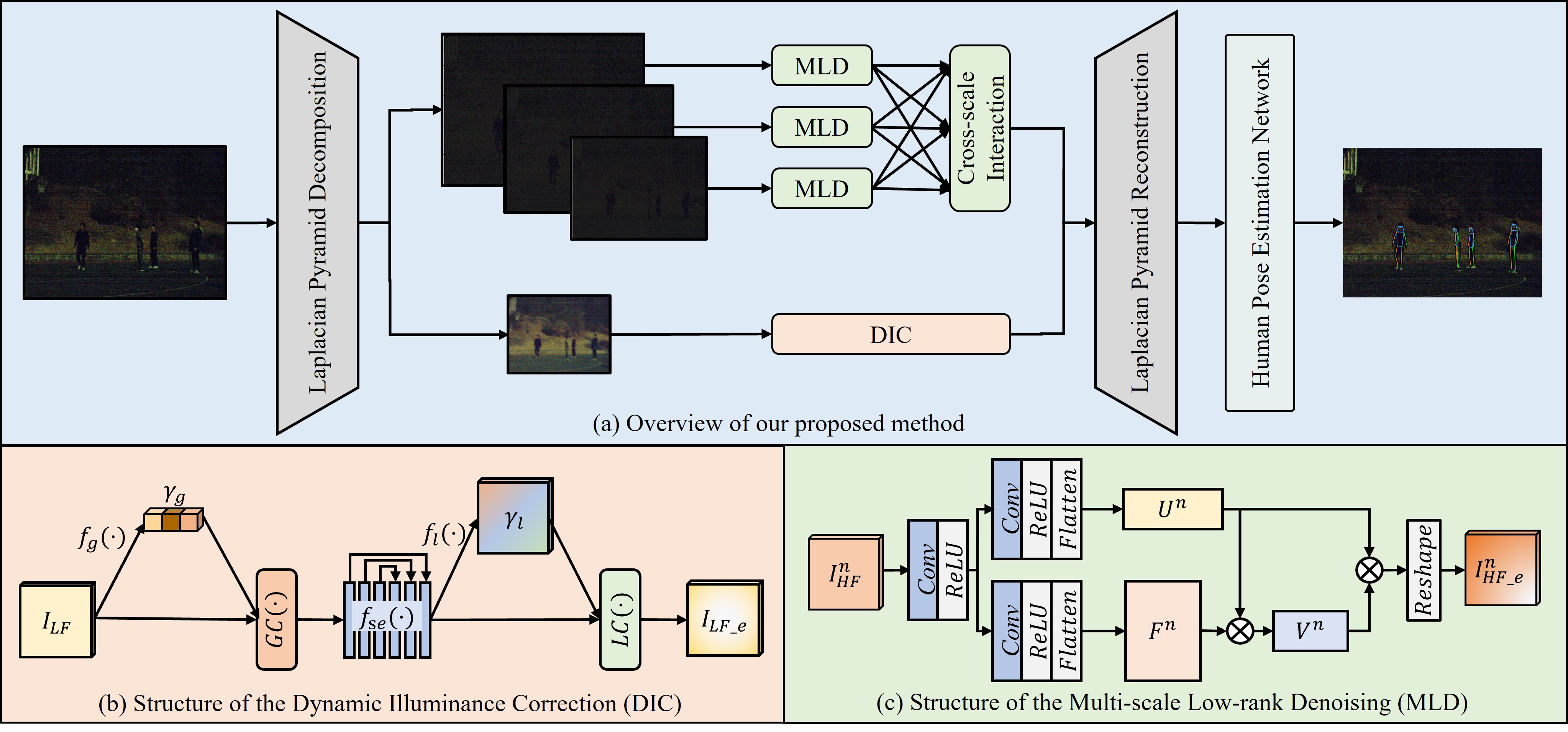}}
\caption{Framework of our proposed methods. We offer a seamless integration of a plug-and-play low-light enhancement module into human pose estimation networks. (a) is the overview of our methods. (b) gives the structure of the dynamic illuminance correction. $GC(\cdot)$ denotes the global correction, and $LC(\cdot)$ represents the local correction. (c) presents the structure of the multi-scale low-rank denoising.
% Employing frequency-domain decomposition, we selectively denoise and enhance high-frequency components through Multi-scale Low-rank Denoising (MLD), while adjusting illuminance of the low-frequency component through Dynamically Illuminance Correction (DIC). The resulting enhanced frequency-domain representations serve as advantageous input images for the subsequent human pose estimation task.
}
\label{model}
\end{figure*}
\vs

As illustrated in Fig.~\ref{model}(a), our method consists of three stages: decomposition, enhancement, and reconstruction. 
In the first stage, we employ a hierarchical Laplacian pyramid to reversibly decompose the image into multi-scale high-frequency components and and a low-frequency base.  
In the low-frequency path of the second stage, dynamic illuminance correction is conducted to enhance semantic clarity.
Leveraging the low-rank structure inherent in natural high-frequency content, we conduct multi-scale low-rank denoising to refine texture details. 
Finally, in the reconstruction stage, the enhanced high- and low-frequency components are recombined to produce a robust and superior representation optimized for human pose estimation.

\subsection{Dynamic illuminance correction}

This unit is designed to achieve efficient dynamic illumination correction to enhance semantic information. To ensure both effectiveness and efficiency, we transform the correction process into a curve coefficients estimation task for global and local illumination adjustment. 
Furthermore, we devised an computationally stable and efficient gamma correction by approximating it with Taylor series expansion.

The structure of the adaptive illumination correction unit is illustrated in Fig.~\ref{model}(c). 
Initially, for the input low-frequency component $I_{LF} \in \mathbf{R}^{ \frac{H}{N}\times \frac{W}{N} \times 3}$, where $H$ and $W$ represent the height and width of the input image, and $N$ denotes the number of decoupling layers, we first compute its global average illumination level. 
Subsequently, a set of global correction coefficients $\gamma_g \in \mathbf{R}^3$ is derived through the function$f_g(\cdot)$:
\begin{equation}
    \gamma_g=f_g(I_{LF})=\sigma(Linear(AvgPool(Conv(I_{LF})))),
\end{equation}
where $Conv(\cdot)$ represents a $3 \times 3$ convolution layer that expands $I_{LF}$ channels to 24, thereby capturing more illumination patterns. $AvgPool(\cdot)$ denotes global average pooling, which reduces spatial dimensions and extracts global statistics. $Linear(\cdot)$ signifies a fully connected layer that captures interrelated information, compressing the dimension to 3. The sigmoid function $\sigma(\cdot)$ then converts the output into global illumination correction coefficients.
We then apply our modified gamma correction formula, utilizing a Taylor series expansion to adjust the overall illumination level of the low-frequency component:
\begin{equation}
    I_{LF}' = I_{LF} \cdot \left(\mathbf{1} + \gamma_g \cdot ln(I_{LF}) + \frac{\gamma_g^2}{2!} \cdot ln^2(I_{LF})\right),
\end{equation}
where the logarithmic function $\ln(\cdot)$ is replaced with the tanh activation function, yielding more stable and smoother adjustment.

Following global correction, a symmetric residual enhancement network $f_{se}(\cdot)$, comprising six convolutional layers, learns semantic enhancements. Given that the low-resolution scale of $I_{LF}$ already represents a compact semantic representation, we employ $3 \times 3$ convolutions instead of larger kernels.

To address the substantial variations in pixel-level lighting conditions in low-light images, we introduce local correction coefficients $\gamma_l$ through the function $f_l(\cdot)$, which aligns with the spatial dimensions of the enhanced image:
\begin{equation}
    \gamma_l = f_l(I'_L) = \sigma(Conv(I'_{LF})).
\end{equation}
This local illumination correction leverages a similar Taylor series expansion for computational efficiency:
\begin{equation}
    I_{LF\_e} = I_{LF}' \cdot \left(\mathbf{1} + \gamma_l \cdot ln(I_{LF}') + \frac{\gamma_l^2}{2!} \cdot ln^2(I_{LF}')\right).
\end{equation}

\subsection{Multi-scale low-rank denoising}

To achieve texture denoising and enhancement, we have devised a denoising module based on the low-rank matrix approximation, complemented by multi-scale interactions to recover details lost during the denoising process.

Initially, we conceptualize the denoising process as a low-rank matrix approximation problem. Specifically, a feature matrix $F_{HF} \in \mathbb{R}^{h \times w \times C}$ derived from high-frequency components is decomposed into the product of two lower-dimensional matrices $U \in \mathbb{R}^{hw \times c}$ and $V \in \mathbb{R}^{c \times C}$, where $c \ll C$. We leverage this inherent redundancy to effectively distinguish the key low-rank structure from the noise:
\begin{equation}
F_{HF} \approx UV.
\end{equation}
% As illustrated in Fig.~\ref{hf}, 
As illustrated in Fig.~\ref{model}(b), 
we employ a $3 \times 3$ convolutional layer to map the $n$-th high-frequency components $I_{HF}^n \in \mathbf{R}^{\frac{H}{n}\times \frac{W}{n} \times 3}$ into a high-dimensional feature space ($C=24$):
\begin{equation}
F_{HF}^n = ReLU(Conv(I_{HF}^n)).
\end{equation}
Subsequently, we utilize convolutional operations to obtain $U^n \in \mathbf{R}^{ \frac{HW}{n^2} \times 3}$, simulating the decomposition process in a differentiable manner to enable end-to-end optimization. The transformation is applied as follows:
\begin{equation}
U^n = Flatten(ReLU(Conv(F_{HF}^n))).
\end{equation}
Concurrently, in a parallel branch, we derive the feature matrix $F^n \in \mathbf{R}^{ \frac{HW}{n^2} \times C}$ via a similar structure:
\begin{equation}
F^n = Flatten(ReLU(Conv(F_{HF}^n))).
\end{equation} 
The feature matrix $F^n$ is then multiplied with the $U^n$ matrix to obtain the $V^n \in \mathbf{R}^{ C \times 3}$ matrix:
\begin{equation}
V^n = F^n \cdot U^n.
\end{equation}
The denoised features are then obtained by multiplying the $U^n$ and $V^n$ matrices, followed by a convolution layer and reshape operation:
\begin{equation}
I_{HF\_e}^n = Reshape(Conv(U^n \cdot V^n)).
\end{equation}

To mitigate the potential loss of fine details during denoising, we leverage the spatial correlations between multi-scale high-frequency components. A cross-scale interaction strategy is implemented post low-rank denoising. This approach fuses the denoised high-frequency components across different scales through simple convolutional layers. This multi-scale interaction not only recovers suppressed details but also enhances the texture quality of the final image output, effectively balancing denoising and texture enhancement.

\section{Experiments}

\subsection{Dataset}

The ExLPose dataset \cite{lee2023human} is tailored for low-light human pose estimation, offering 2,556 pairs of real low-light and well-lit images with precise human body pose annotations. Captured across 251 scenes using a dual-camera setup, it includes 14,215 instances. The dataset is divided into a training set of 2,065 pairs and a test set of 491 pairs. Annotations follow the CrowdPose format, marking 14 key human joints for pose estimation. We evaluate model performance using the mean Average Precision (mAP) score based on Object Keypoint Similarity (OKS). The ExLPose dataset also offers three difficulty-based test subsets: Low-Light Normal (\textbf{LL-N}), Low-Light Hard (\textbf{LL-H}), and Low-Light Extreme (\textbf{LL-E}), along with an aggregate \textbf{LL-A} and a Well-Lit (\textbf{WL}) subset for comprehensive evaluation.

\subsection{Implementation details}
All experiments were conducted on an RTX 3090 GPU with 24GB of memory. On the ExLPose dataset, all models were trained with a batch size of 32, using a learning rate of 5e-4 and the Adam optimizer \cite{kingma2014adam}, over 210 epochs. We employed ResNet-50 \cite{he2016deep} as our human pose estimation network for simplicity. For comparison methods, we utilized their official implementations with default optimal settings to ensure fair evaluation. Low-light image enhancement methods were retrained on the ExLPose dataset, and the well-trained models were used to enhance low-light images. For methods aimed at learning illumination-invariant features, well-lit images were considered the source domain, while low-light images served as the target domain.

\subsection{Main results}

\begin{table}[htbp]
\caption{Comparisons with state-of-the-art methods on the ExLPose dataset.}
\begin{center}
\resizebox{0.8\linewidth}{!}{ % resizebox
\begin{tabular}{|c|c|c|c|c|c|}
\hline
\textbf{Method} & \textbf{LL-N} & \textbf{LL-H} & \textbf{LL-E} & \textbf{LL-A} & \textbf{WL}  \\
\hline
\multicolumn{6}{|c|}{\textbf{\textit{Restoration-based methods}}}\\
\hline
LIME  \cite{guo2016lime} & 38.3 & 25.6 & 12.5 & 26.6 & 63.0  \\
Zero-DCE \cite{guo2020zero} & 51.0 & 40.1 & 24.6 & 39.7 & 76.2  \\
RUAS \cite{liu2021retinex} & \underline{52.5} & \underline{41.4} & \underline{24.7} & \underline{40.7} & 76.7  \\
\hline
\multicolumn{6}{|c|}{\textbf{\textit{Adaptation-based methods}}}\\
\hline
DANN \cite{ganin2016domain} & 34.9 & 24.9 & 13.3 & 25.4 & 58.6  \\
AdvEnt \cite{vu2019advent} & 35.6 & 23.5 & 8.8 & 23.8 & 62.4  \\
ExLPose \cite{lee2023human} & 42.3 & 34.0 & 18.6 & 32.7 & 68.5  \\
\hline
\multicolumn{6}{|c|}{\textbf{\textit{Enhancement-based methods}}}\\
\hline
PENet \cite{yin2023pe} & 46.2 & 36.3 & 23.1 & 36.3 & 73.7  \\
DENet \cite{qin2022denet} & 51.8 & 41.2 & 24.1 & 40.3 & 76.3  \\
FeatEnHancer \cite{hashmi2023featenhancer} & 50.3 & 36.7 & 20.4 & 37.0 & \textbf{77.2} \\
Ours & \textbf{53.7} & \textbf{42.0} & \textbf{25.4} & \textbf{41.7} & \underline{76.8}  \\
\hline
\end{tabular}
}% resizebox
\end{center}
\label{tab:main}
\end{table}
% \vs

As shown in Table~\ref{tab:main}, we classify state-of-the-art methods into three distinct paradigms: restoration-based methods use low-light enhancement as preprocessing, adaptation-based methods learn illumination-invariant features, and enhancement-based methods employ task-driven, end-to-end training. Notably, our method consistently outperforms these approaches across all three categories in low-light conditions, while remaining highly competitive in well-lit environments. 
Restoration-based methods, despite good performance, require pre-training and additional processing for low-light data, resulting in significant time overhead. In contrast, our approach achieves better results with minimal extra parameters and no extra processing delay.
Due to the substantial domain gap between extremely low-light and well-lit images, adaptation-based methods struggle to bridge this gap, resulting in suboptimal performance, and their reliance on paired training data further limits practical application. Our method, however, does not require paired data and demonstrates robust performance across different lighting conditions.
Although enhancement-based methods are efficient and simple to train, their performance is hindered under extreme low-light conditions, particularly methods like FeatEnhancer that focus on enhancing semantic information. In contrast, our method handles varying lighting conditions more effectively.

\subsection{Ablation study}

\textbf{Effect of DIC and MLD.}
As shown in the Table~\ref{tab:module}, we validated the effectiveness of the two proposed unit and their impact on network parameters and computational cost. The results indicate that adding dynamic illumination correction (DIC) improves accuracy on the LL-A subset by 4.4\%, while multi-scale low-rank denoising (MLD) in the high-frequency pathway alone boosts performance by 5.1\%. When both modules are applied, our proposed approach achieves a 5.8\% improvement, further demonstrating the effectiveness of the model structure. 
Additionally, performance gains on the WL subset confirm the generalizability of our method, which focuses on enhancing task-relevant information.
Regarding the additional computational cost, using DIC alone introduces only 0.04M parameters and 0.04GFLOPs, while MLD adds 0.07M parameters and 1.33GFLOPs. The complete module, when applied, results in a modest increase of 0.11M parameters and 1.36GFLOPs.

\begin{table}[htbp]
\centering
\caption{Effect of Dynamic Illuminance Correction (DIC) and Multi-scale Low-rank Denoising (MLD) on the ExLPose dataset.}
\resizebox{0.8\linewidth}{!}{
\begin{tabular}{|c|c|c|c|c|c|}
\hline
\textbf{DIC} & \textbf{MLD} & \textbf{LL-A} & \textbf{WL} & \textbf{Param(M)} & \textbf{FLOPs(G)} \\
\hline
& & 35.9 & 72.3 & 34.0 & 5.45 \\
\checkmark & & 40.3 & 76.6 & 34.04 & 5.49 \\
& \checkmark & 41.0 & 77.0 & 34.07 & 6.78 \\
\checkmark & \checkmark & 41.7 & 76.8 & 34.11 & 6.81 \\
\hline
\end{tabular}
}% resizebox
\label{tab:module}
\end{table}
% \vs

\textbf{Effect of the order of global and local corrections in DIC.}
we compared with an alternative design -- the local-to-global order, the opposite direction of what we propose. 
We conduct an ablation on the ExLPose dataset. 
As shown in Table~\ref{tab:dic}, we observe that our global-to-local design clearly outperforms the other, validating the coarse-to-fine idea behind our design - global correction should come first to adjust the overall illumination, facilitating the subsequent more detailed local refinements, and finally ensuring more robust and effective enhancement.
\begin{table}[h!]
% \vspace{-0.5em}
\centering
\caption{Ablation on the design of DIC module.}
\label{tab:dic}
\resizebox{0.8\linewidth}{!}{ % reizebox
\begin{tabular}{|c|c|c|c|c|c|}
\hline
 Order          & LL-N & LL-H & LL-E & LL-A & WL   \\ \hline
global-to-local & 52.1 & 40.9 & 24.2 & 40.6 & 75.8 \\ \hline
local-to-global & 50.9 & 39.7 & 23.2 & 39.1 & 76.9 \\ \hline
\end{tabular}
}% resizebox
% \vspace{-1.0em}
\end{table}
% \vs

\textbf{Effect of the number of levels of Laplacian pyramid.}

We evaluated the impact of the number of Laplacian decomposition layers on the framework's performance. As shown in Table~\ref{tab:layer}, optimal performance is achieved with a four-layer decomposition, comprising three high-frequency components and one low-frequency component. Increasing the number of layers beyond this configuration leads to a gradual decline in performance. Therefore, we adopt the four-layer Laplacian pyramid as the default optimal setting.

\begin{table}[htbp]
\centering
\caption{Effect of the number of levels of Laplacian pyramid.}
\setlength{\tabcolsep}{3mm}
\resizebox{0.8\linewidth}{!}{ % resizebox
\begin{tabular}{|c|c|c|c|c|}
\hline
\textbf{Number} & \textbf{LL-A} & \textbf{WL} & \textbf{Param(M)} & \textbf{FLOPs(G)} \\ 
\hline
0   & 35.9 & 72.3 & 34.00 & 5.45 \\
2   & 40.8 & 76.6 & 34.06 & 6.81\\
3   & 41.3 & 76.2 & 34.08 &6.74 \\
4   & 41.7 & 76.8 & 34.11 &6.81 \\
5   & 41.1 & 76.3 & 34.17 &6.91 \\
6   & 40.7 & 76.3 & 34.24 & 7.02\\  
\hline
\end{tabular}
}% resizebox
\label{tab:layer}
\end{table}
% \vs

\textbf{Visual comparisons with state-of-the-art methods.}
we presented visual comparisons across varying low-light conditions across a range of difficulties (e.g., normal, hard, and extremely hard low-light cases), with proper zooming in for better visualization. 
To facilitate visual comparison, the corresponding well-lit images will be also provided, as shown in Fig.~\ref{fig:visual_comparison} below. 
Consistent with our quantitative evaluation, these examples also highlight the performance advantage of our method, achieving more accurate predictions even in extremely low-light conditions with severe noises.
In contrast, previous state-of-the-art methods often fail, producing incorrect keypoints or even implausible poses.
\begin{figure}[h!]
\centering
{\includegraphics[width=0.45\textwidth]{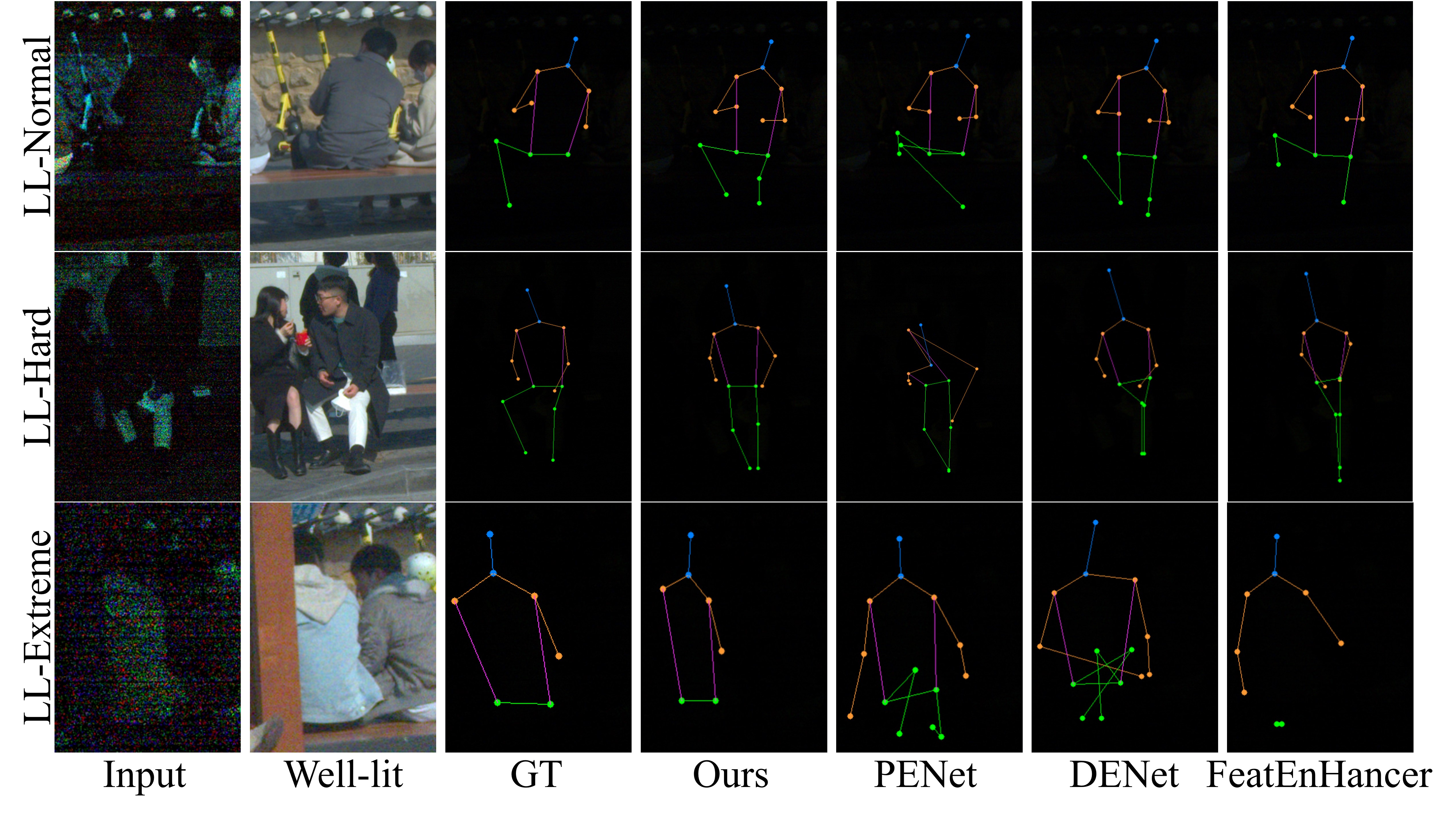}}
\caption{Visual comparison of human pose estimation results among our method and current state of the art approaches (PENet \cite{yin2023pe}, DENet \cite{qin2022denet}, FeatEnHancer \cite{hashmi2023featenhancer}).}
\label{fig:visual_comparison}
% \vspace{-0.5em}
\end{figure}

\section{Conclusion}

In this work, we propose an efficient method for low-light human pose estimation. Our frequency-decoupling framework applies dynamic illumination correction to low-frequency components to better enhance semantics, and leverages the low-rank properties of high-frequency components for denoising to improve texture preservation. Extensive experiments demonstrate that our method outperforms the current state-of-the-art approaches.

\section*{Acknowledgment}

This work is supported by the National Natural Science Foundation of China (Grant No. 62202241), Jiangsu Province Natural Science Foundation for Young Scholars (Grant No. BK20210586), NUPTSF (Grant No. NY221018) and Double-Innovation Doctor Program under Grant JSSCBS20220657.

\bibliographystyle{IEEEbib}
\bibliography{ref}

\begin{thebibliography}{10}

\bibitem{zanfir2023hum3dil}
Andrei Zanfir, Mihai Zanfir, Alex Gorban, Jingwei Ji, Yin Zhou, Dragomir Anguelov, and Cristian Sminchisescu,
\newblock ``Hum3dil: Semi-supervised multi-modal 3d humanpose estimation for autonomous driving,''
\newblock in {\em Conference on Robot Learning}. PMLR, 2023, pp. 1114--1124, {PMLR}.

\bibitem{sengupta2020mm}
Arindam Sengupta, Feng Jin, Renyuan Zhang, and Siyang Cao,
\newblock ``mm-pose: Real-time human skeletal posture estimation using mmwave radars and cnns,''
\newblock {\em IEEE Sensors Journal}, vol. 20, no. 17, pp. 10032--10044, 2020.

\bibitem{cucchiara2022fine}
Rita Cucchiara and Matteo Fabbri,
\newblock ``Fine-grained human analysis under occlusions and perspective constraints in multimedia surveillance,''
\newblock {\em ACM Transactions on Multimedia Computing, Communications, and Applications (TOMM)}, vol. 18, no. 1s, pp. 1--23, 2022.

\bibitem{cormier2022we}
Mickael Cormier, Aris Clepe, Andreas Specker, and J{\"u}rgen Beyerer,
\newblock ``Where are we with human pose estimation in real-world surveillance?,''
\newblock in {\em Proceedings of the IEEE/CVF Winter Conference on Applications of Computer Vision}. 2022, pp. 591--601, {IEEE}.

\bibitem{toshev2014deeppose}
Alexander Toshev and Christian Szegedy,
\newblock ``Deeppose: Human pose estimation via deep neural networks,''
\newblock in {\em Proceedings of the IEEE conference on computer vision and pattern recognition}, 2014, pp. 1653--1660.

\bibitem{li2021human}
Jiefeng Li, Siyuan Bian, Ailing Zeng, Can Wang, Bo~Pang, Wentao Liu, and Cewu Lu,
\newblock ``Human pose regression with residual log-likelihood estimation,''
\newblock in {\em Proceedings of the IEEE/CVF International Conference on Computer Vision}, 2021, pp. 11025--11034.

\bibitem{sun2019deep}
Ke~Sun, Bin Xiao, Dong Liu, and Jingdong Wang,
\newblock ``Deep high-resolution representation learning for human pose estimation,''
\newblock in {\em Proceedings of the IEEE/CVF conference on computer vision and pattern recognition}, 2019, pp. 5693--5703.

\bibitem{guo2016lime}
Xiaojie Guo, Yu~Li, and Haibin Ling,
\newblock ``Lime: Low-light image enhancement via illumination map estimation,''
\newblock {\em IEEE Transactions on image processing}, vol. 26, no. 2, pp. 982--993, 2016.

\bibitem{jiang2021enlightengan}
Yifan Jiang, Xinyu Gong, Ding Liu, Yu~Cheng, Chen Fang, Xiaohui Shen, Jianchao Yang, Pan Zhou, and Zhangyang Wang,
\newblock ``Enlightengan: Deep light enhancement without paired supervision,''
\newblock {\em IEEE transactions on image processing}, vol. 30, pp. 2340--2349, 2021.

\bibitem{guo2020zero}
Chunle Guo, Chongyi Li, Jichang Guo, Chen~Change Loy, Junhui Hou, Sam Kwong, and Runmin Cong,
\newblock ``Zero-reference deep curve estimation for low-light image enhancement,''
\newblock in {\em Proceedings of the IEEE/CVF conference on computer vision and pattern recognition}, 2020, pp. 1780--1789.

\bibitem{liu2021retinex}
Risheng Liu, Long Ma, Jiaao Zhang, Xin Fan, and Zhongxuan Luo,
\newblock ``Retinex-inspired unrolling with cooperative prior architecture search for low-light image enhancement,''
\newblock in {\em Proceedings of the IEEE/CVF conference on computer vision and pattern recognition}, 2021, pp. 10561--10570.

\bibitem{lee2023human}
Sohyun Lee, Jaesung Rim, Boseung Jeong, Geonu Kim, Byungju Woo, Haechan Lee, Sunghyun Cho, and Suha Kwak,
\newblock ``Human pose estimation in extremely low-light conditions,''
\newblock in {\em Proceedings of the IEEE/CVF conference on computer vision and pattern recognition}, 2023, pp. 704--714.

\bibitem{cui2021multitask}
Ziteng Cui, Guo-Jun Qi, Lin Gu, Shaodi You, Zenghui Zhang, and Tatsuya Harada,
\newblock ``Multitask aet with orthogonal tangent regularity for dark object detection,''
\newblock in {\em Proceedings of the IEEE/CVF international conference on computer vision}, 2021, pp. 2553--2562.

\bibitem{hashmi2023featenhancer}
Khurram~Azeem Hashmi, Goutham Kallempudi, Didier Stricker, and Muhammad~Zeshan Afzal,
\newblock ``Featenhancer: Enhancing hierarchical features for object detection and beyond under low-light vision,''
\newblock in {\em Proceedings of the IEEE/CVF International Conference on Computer Vision}, 2023, pp. 6725--6735.

\bibitem{schiller2010parallel}
Peter~H Schiller,
\newblock ``Parallel information processing channels created in the retina,''
\newblock {\em Proceedings of the National Academy of Sciences}, vol. 107, no. 40, pp. 17087--17094, 2010.

\bibitem{markovsky2012low}
Ivan Markovsky,
\newblock {\em Low rank approximation: algorithms, implementation, applications}, vol. 906,
\newblock Springer, 2012.

\bibitem{kingma2014adam}
Diederik~P. Kingma and Jimmy Ba,
\newblock ``Adam: {A} method for stochastic optimization,''
\newblock in {\em 3rd International Conference on Learning Representations, {ICLR} 2015, San Diego, CA, USA, May 7-9, 2015, Conference Track Proceedings}, Yoshua Bengio and Yann LeCun, Eds., 2015.

\bibitem{he2016deep}
Kaiming He, Xiangyu Zhang, Shaoqing Ren, and Jian Sun,
\newblock ``Deep residual learning for image recognition,''
\newblock in {\em Proceedings of the IEEE conference on computer vision and pattern recognition}, 2016, pp. 770--778.

\bibitem{ganin2016domain}
Yaroslav Ganin, Evgeniya Ustinova, Hana Ajakan, Pascal Germain, Hugo Larochelle, Fran{\c{c}}ois Laviolette, Mario March, and Victor Lempitsky,
\newblock ``Domain-adversarial training of neural networks,''
\newblock {\em Journal of machine learning research}, vol. 17, no. 59, pp. 1--35, 2016.

\bibitem{vu2019advent}
Tuan-Hung Vu, Himalaya Jain, Maxime Bucher, Matthieu Cord, and Patrick P{\'e}rez,
\newblock ``Advent: Adversarial entropy minimization for domain adaptation in semantic segmentation,''
\newblock in {\em Proceedings of the IEEE/CVF conference on computer vision and pattern recognition}. 2019, pp. 2517--2526, Computer Vision Foundation / {IEEE}.

\bibitem{yin2023pe}
Xiangchen Yin, Zhenda Yu, Zetao Fei, Wenjun Lv, and Xin Gao,
\newblock ``Pe-yolo: Pyramid enhancement network for dark object detection,''
\newblock in {\em International Conference on Artificial Neural Networks}. Springer, 2023, pp. 163--174.

\bibitem{qin2022denet}
Qingpao Qin, Kan Chang, Mengyuan Huang, and Guiqing Li,
\newblock ``Denet: detection-driven enhancement network for object detection under adverse weather conditions,''
\newblock in {\em Proceedings of the Asian Conference on Computer Vision}. 2022, pp. 2813--2829, Springer.

\end{thebibliography}

\end{document}